\documentclass[twocolumn]{article}
\usepackage{graphicx} 
\usepackage[round]{natbib}
\usepackage{url}
\usepackage{booktabs}
\usepackage{booktabs, tabularx, multirow, threeparttable, xcolor}
\usepackage{tabularx,booktabs,array}
\newcolumntype{Y}{>{\raggedright\arraybackslash}X}
\newcolumntype{L}[1]{>{\raggedright\arraybackslash}p{#1}}

\title{Vision-Grounded Machine Interpreting: Improving the Translation Process through Visual Cues}
\author{Claudio Fantinuoli\\
University of Mainz\\
\texttt{fantinuoli@uni-mainz.de}
}
\date{} 

\begin{document}

\newcommand{\Cone}{\textbf{C1}:~Speech-only}
\newcommand{\Ctwo}{\textbf{C2}:~Speech+Scene Description}
\newcommand{\Cthree}{\textbf{C3}:~Direct Multimodal}
\newcommand{\Cfour}{\textbf{C4}:~Adversarial}

\maketitle

\begin{abstract}
Machine Interpreting systems are currently implemented as unimodal, real-time speech-to-speech architectures, processing translation exclusively on the basis of the linguistic signal. Such reliance on a single modality, however, constrains performance in contexts where disambiguation and adequacy depend on additional cues, such as visual, situational, or pragmatic information. This paper introduces Vision-Grounded Interpreting (VGI), a novel approach designed to address the limitations of unimodal machine interpreting. We present a prototype system that integrates a vision–language model to process both speech and visual input from a webcam, with the aim of priming the translation process through contextual visual information. To evaluate the effectiveness of this approach, we constructed a hand-crafted diagnostic corpus targeting three types of ambiguity. In our evaluation, visual grounding substantially improves lexical disambiguation, yields modest and less stable gains for gender resolution, and shows no benefit for syntactic ambiguities. We argue that embracing multimodality represents a necessary step forward for advancing translation quality in machine interpreting. 
\end{abstract}

\section{Introduction}

Machine Interpreting, the live translation of spoken language in dynamic communicative settings, is most commonly implemented as a unimodal pipeline, combining automatic speech recognition, machine translation, and speech synthesis. More recently, end-to-end architectures have emerged as an alternative to this cascaded approach \citep{sperberSpeechTranslationEndtoEnd2020,fantinuoliMachineInterpreting2025, papi_direct_2023}. Despite significant advances, current architectures continue to rely predominantly on the linguistic signal for meaning inference and output generation. End-to-end models extend this paradigm by leveraging acoustic features such as prosody, stress, and intonation as additional cues for disambiguation \citep{tsiamasSpeechMoreWords2024,seamlesscommunicationteamJointSpeechText2025}, yet the limitations of unimodal approaches remain evident. In particular, existing systems struggle when successful interpretation depends on information beyond the spoken channel. Ambiguities involving gender resolution, deixis, lexical polysemy, and syntactic structure frequently require extralinguistic grounding that speech alone cannot provide \citep{berzakYouSeeWhat2016,pavoneResolvingLinguisticAmbiguities2022}.

Human communication and interpreting, by contrast, are intrinsically multimodal \citep{gileBasicConceptsModels2009,pochhackerIntroducingInterpretingStudies2016,pochhackerMachineInterpretingInterpreting2024}. Face-to-face interaction unfolds within a shared environment saturated with contextual cues, including facial expressions, gestures, and situational information. Professional interpreters routinely exploit these resources to resolve ambiguity and adapt to communicative demands \citep{arbonaRoleSemanticallyRelated2024,leontevaMetaphoricGesturesSimultaneous2023,arbonaSemanticallyRelatedGestures2023}. By contrast, current real-time speech translation systems remain largely insensitive to such signals, leading to degraded performance and translations that are insufficiently anchored in the communicative event.

In this paper, we introduce the concept of Vision-Grounded Interpreting (VGI) as a means of addressing these limitations. Building on studies of multimodality in other areas of translation, VGI extends the classical speech-to-speech translation pipeline with a visual component, either  (i) by employing a vision–language model that generates live scene descriptions from webcam input and integrates them with the source speech at inference time, or (ii) by deploying a multimodal model capable of jointly processing speech and visual input. The information retrieved by the vision system functions as a compact symbolic proxy for situational grounding: it aims to identify participants, objects, and activities in the environment. This information can then be used to condition the translation choices made by the model, reducing ambiguities and producing more contextually appropriate lexical and syntactic choices \citep{radfordLearningTransferableVisual, alayracFlamingoVisualLanguage2022}.


\begin{figure}[t]
\centering
\includegraphics[width=0.9\linewidth]{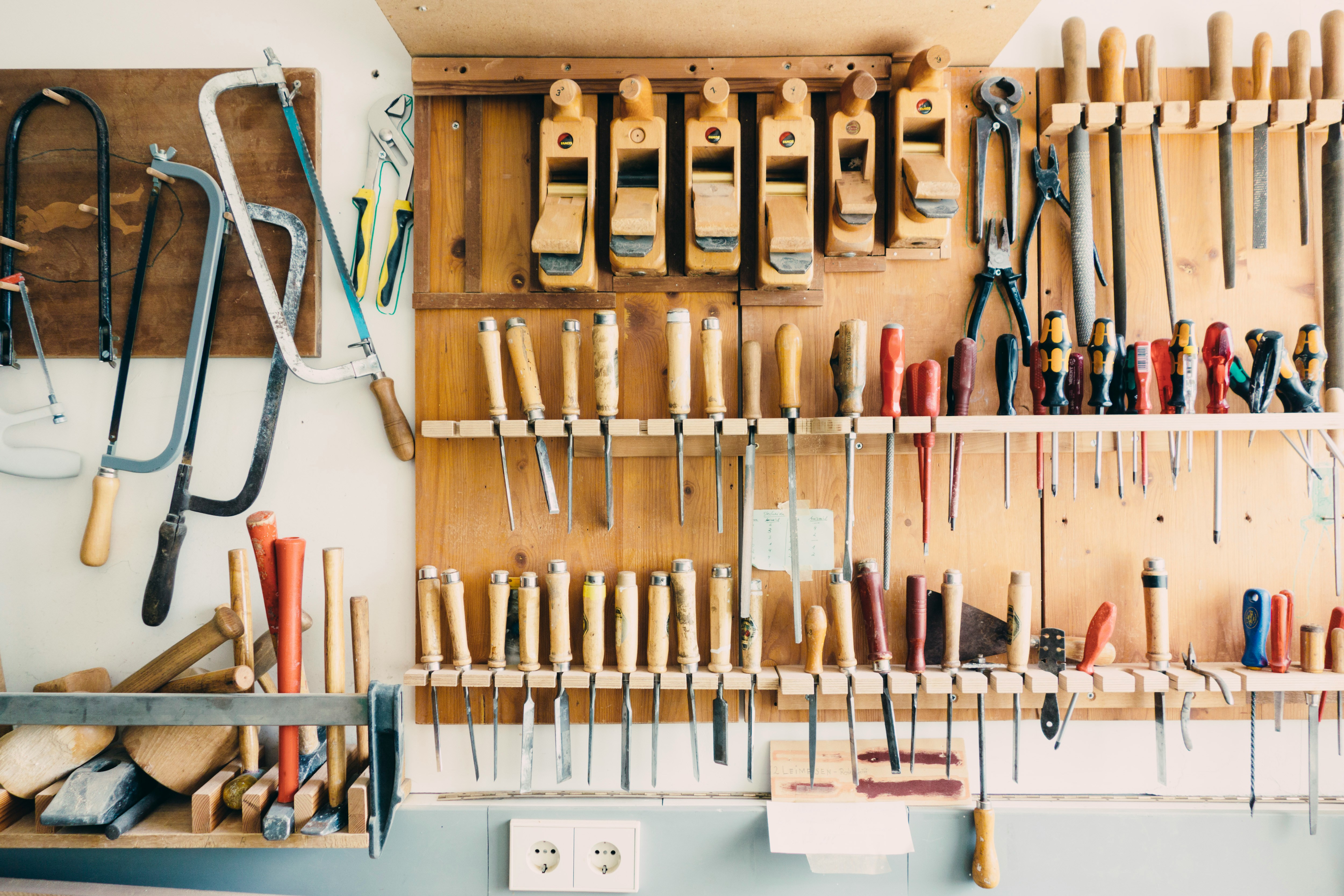}
\caption{Example image used in the corpus to condition the translation of the ambiguous word \textit{chiave}. }
\label{fig:example_corpus}
\end{figure}

To illustrate the expectations for a VGI system, consider Figure~\ref{fig:example_corpus}. A speaker utters the Italian sentence ``Passami la chiave'' (``Give me the key/wrench''), in which the noun \textit{chiave} is lexically ambiguous. In this instance, the intended meaning ``wrench'' cannot be derived from the linguistic signal alone. Incorporating visual cues into the translation process is expected to resolve such ambiguity: for example, by detecting that the speaker is pointing toward a wrench, or by recognizing that the surrounding environment, such as a workshop in our example, renders the technical interpretation more plausible.

To evaluate this approach and identify the most suitable strategy for our prototype, we developed a small diagnostic corpus of hand-crafted problem triggers targeting three categories of weakness in unimodal systems: lexical disambiguation, gender resolution, and syntactic disambiguation. We compared performance across four experimental conditions (\Cone{}, \Ctwo{}, \Cthree{}, \Cfour{}). We test this with a naive prompt approach, meaning that we do not elaborate on complex prompt that are supposed to increase gains in quality. This design allows us to assess not only the overall benefit of incorporating visual context, but also the extent of improvement, the relative effectiveness of different integration strategies, and the robustness of the system when confronted with potentially misleading information.

Beyond controlled evaluation, our prototype was also tested in authentic communicative settings, using the integration strategy that yielded the best performance in the diagnostic experiments. These tests serve to demonstrate not only the feasibility of multimodal extensions under experimental conditions, but also their practical viability in realistic deployment scenarios.

\section{Related Work}

Research relevant to VGI has progressed along three main strands: (i) extending multimodality to speech translation, (ii) developing large multimodal language models (MLLMs) with the ability to process visual input, and (iii) addressing ambiguity and bias in machine translation (MT).

In multimodal speech translation, recent advances have aimed at unifying speech and text within single, direct models. For example, Meta’s SeamlessM4T \citep{seamlessm4t} integrates both modalities across more than 100 languages, while AV2AV frameworks \citep{choiAV2AVDirectAudioVisual2024} extend this paradigm to audio–visual speech-to-speech translation, showing benefits in noisy conditions and for lip-synchronized output \citep{liuDetectDisambiguateTranslate}. Complementary approaches drawing on information-theoretic and error-aware methods further demonstrate that visual cues can compensate for imperfect transcripts \citep{ji2022visual,li2021visual}.

Parallel efforts have focused on building multimodal language models capable of processing text and vision jointly, thereby enhancing the reasoning capacity of large language models. The VCR dataset \citep{zellers2019vcr} introduced the idea that visual understanding requires commonsense reasoning beyond object recognition. Technically, most MLLMs connect a vision encoder to an LLM through an adapter: LLaVA \citep{liu2023llava} applies visual instruction tuning, BLIP-2 \citep{li2023blip2} introduces the Q-Former for efficient alignment, and InternVL \citep{chen2024internvl,chen2025internvl} scales the visual encoder for higher accuracy. For dynamic scenes, models such as F-16 \citep{li2025f16} process video at 16 FPS to capture fine-grained motion, outperforming sparse-frame approaches.

Within MT, it is well established that systems frequently struggle with gender resolution, deixis, and lexical polysemy. Persistent gender bias has been documented across benchmarks \citep{stanovsky2019gender,cho2023gender}, though context-aware MT and multimodal extensions can alleviate some cases \citep{corral2024gebnlp}. Various strategies have been explored to integrate visual information into MT, including information-theoretic approaches to increase visual awareness \citep{jiIncreasingVisualAwareness2022}, semantic scene graph representations \citep{hatamiLeveragingVisualScene2025}, visual pivoting for low-resource language pairs \citep{tayirVisualPivotingUnsupervised2024}, and the use of visual cues to overcome noisy textual input \citep{liVisualCuesError2021}.

Taken together, prior work demonstrates why and how visual input can be integrated into translation pipelines, yet most studies have concentrated on written translation or offline speech translation for specific applications such as dubbing. Few have systematically examined \emph{how} different integration strategies perform under controlled ambiguity triggers in speech, or \emph{how} models behave when visual input is misleading. Our study addresses this gap by focusing on real-time machine interpreting, introducing a diagnostic evaluation corpus targeting three specific sources of ambiguity and testing integration strategies tailored to this modality.

\section{System Architecture}
\subsection{Formal description}

To incorporate visual scene information into machine interpreting, we model the AI interpreter as a large language model (LLM) conditioned on both speech-derived text and an auxiliary image describing the current environment in which the utterance has bèn said. Let 
\[
S = (s_{1}, s_{2}, \dots, s_{n})
\] 
denote the source-language utterance, and let 
\[
I = (i_{1}, i_{2}, \dots, i_{m})
\] 
denote the image captured by a vision sytstem. 

In a text-only scenario, the LLM receives a prompt 
\[
\mathcal{P} = f(S),
\]
where \(f\) is a formatting function that encodes the source-language utterance into the model’s context. In the augmented scenario, the prompt is extended as
\[
\mathcal{P}' = f(S, I),
\]
typically realized as concatenation in structured form, e.g. 
\texttt{[Image: $C$] [Source speech: $S$]}. 

During decoding, the model then computes
\[
\hat{T} = \arg\max_{T} P(T \mid \mathcal{P}'),
\]
where \(\hat{T}\) is the target-language hypothesis. By conditioning on \(I\), the model gains access to contextual priors over domains, entities, and actions, which guide lexical disambiguation, anaphora resolution, and pragmatic adaptation. This design aligns with recent advances in multimodal instruction-tuned models \citep{alayracFlamingoVisualLanguage2022,liBLIP2BootstrappingLanguageImage2023,liuVisualInstructionTuning2023}, which demonstrate how caption-like descriptions enable LLMs to perform contextually grounded reasoning. 

\subsection{Prototype}
We implemented a prototype of a simultaneous speech-to-speech translation system following the classical cascaded pipeline of ASR→MT→TTS. A vision component was added to monitor the communicative setting via a webcam. The system supports two integration strategies: (i) a vision-to-caption module, in which images are sampled when significant scene changes are detected and converted into concise textual descriptions that are combined with the speech input; and (ii) direct multimodal processing, in which speech and images are provided jointly to the translation model. 

For evaluation, we used GPT-4o, as it supports both caption generation and direct multimodal integration, while also providing a unified framework suitable for controlled comparison.

\begin{figure}[t]
\centering
\includegraphics[width=0.9\linewidth]{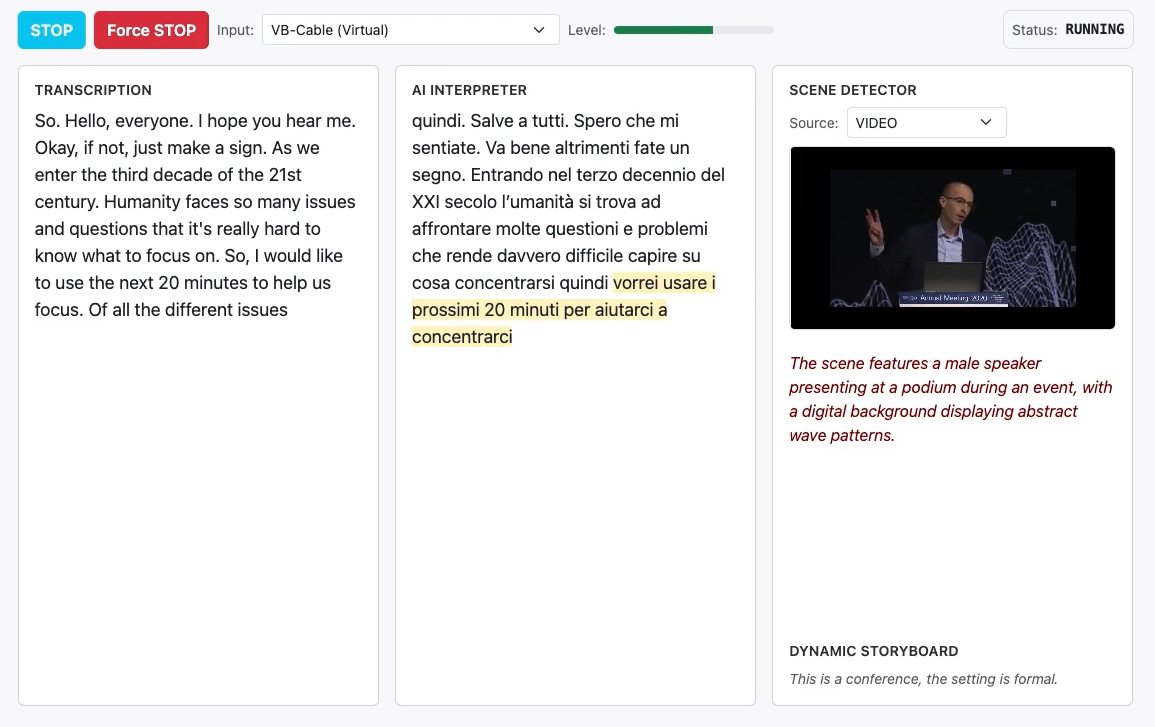}
\caption{Prototype of VGI.}
\label{fig:protoptype}
\end{figure}

\section{Experiment}

\subsection{Method}

We evaluate the system with three categories of ambiguity (lexical disambiguation, gender resolution, and syntactic disambiguation) and four conditions that differ in how visual context is integrated into translation. Besides a \Cone{} baseline, we test two variants combining source and image (\Ctwo{} concatenating source utterance with the caption of the image, and \Cthree{} through multimodal direct ingestion where the model processes image and text jointly), and an \Cfour{} setting with mismatched captions. This design allows us to compare baseline performance, potential gains from visual grounding, and robustness to misleading input.

\begin{table*}[t]
\centering
\caption{Experimental conditions used in the study.}
\label{tab:conditions}
\begin{tabularx}{\textwidth}{@{} L{5cm} Y @{}}
\toprule
\textbf{Condition} & \textbf{Operationalization} \\
\midrule
\Cone   & Translate from the source sentence only (no visual context). Serves as the main baseline. \\
\Ctwo   & Instruction-style prompt: a caption is generated independently and used as context with an explicit policy to resolve ambiguity at translation time. \\
\Cthree & Provide the source sentence and the image simultaneously to a vision-capable model, bypassing intermediate captions. \\
\Cfour  & Combine the source sentence with a mismatched or irrelevant caption from another scene to test robustness against misleading context. \\
\bottomrule
\end{tabularx}
\end{table*}

\subsection{Primary Hypotheses \& Key Comparisons}

To assess the role of visual context in machine interpreting, we formulated hypotheses that connect the experimental conditions to specific types of translation challenges. Our expectation is that adding visual information should be particularly beneficial for triggers such as gender resolution (e.g., \emph{doctor} → \emph{dottore/dottoressa}), lexical polysemy (words with multiple possible senses), and syntactic ambiguity (structural alternatives such as ``Paul bought green shirts and shoes.'', where green can refer to shirts only or to shirts and shoes). Beyond the overall question of whether multimodal input improves translation relative to a source-only baseline, we also test whether different integration strategies (concatenation and direct multimodal input) vary in effectiveness across these triggers, and whether systems remain robust when presented with misleading contextual cues. The following key comparisons operationalize these hypotheses.

\begin{itemize}
  \item \textbf{H1 (visual helps):} C2, C3 $\;>$ C1 on adequacy, contextual fit, and trigger resolution (gender, polysemy, syntax).
  \item \textbf{H2 (caption vs multimodal direct):} C3 $\ge$ C2 (if direct multimodal grounding helps beyond textual captions).
  \item \textbf{H3 (robustness):} C4 does not degrade vs.\ C1 (misleading context should not harm; avoid over-reliance).
\end{itemize}

\subsection{Corpus description}

To evaluate the proposed approach, we constructed a small diagnostic corpus specifically designed to trigger the kinds of ambiguities that unimodal systems often fail to resolve. Unlike large-scale benchmarks for machine translation or speech translation, our goal was not coverage, but targeted evaluation under controlled conditions.

The corpus consists of 120 source utterances paired with 120 images collected via webcam or selected from open repositories. Each utterance was designed or selected to instantiate one of three ambiguity triggers: (i) \emph{lexical polysemy}, (ii) \emph{gender resolution}, and (iii) \emph{syntactic ambiguity}. For each category, we created 40 examples, producing a balanced dataset in which every trigger is represented in multiple contexts.

\begin{figure}[t]
\centering
\includegraphics[width=0.9\linewidth]{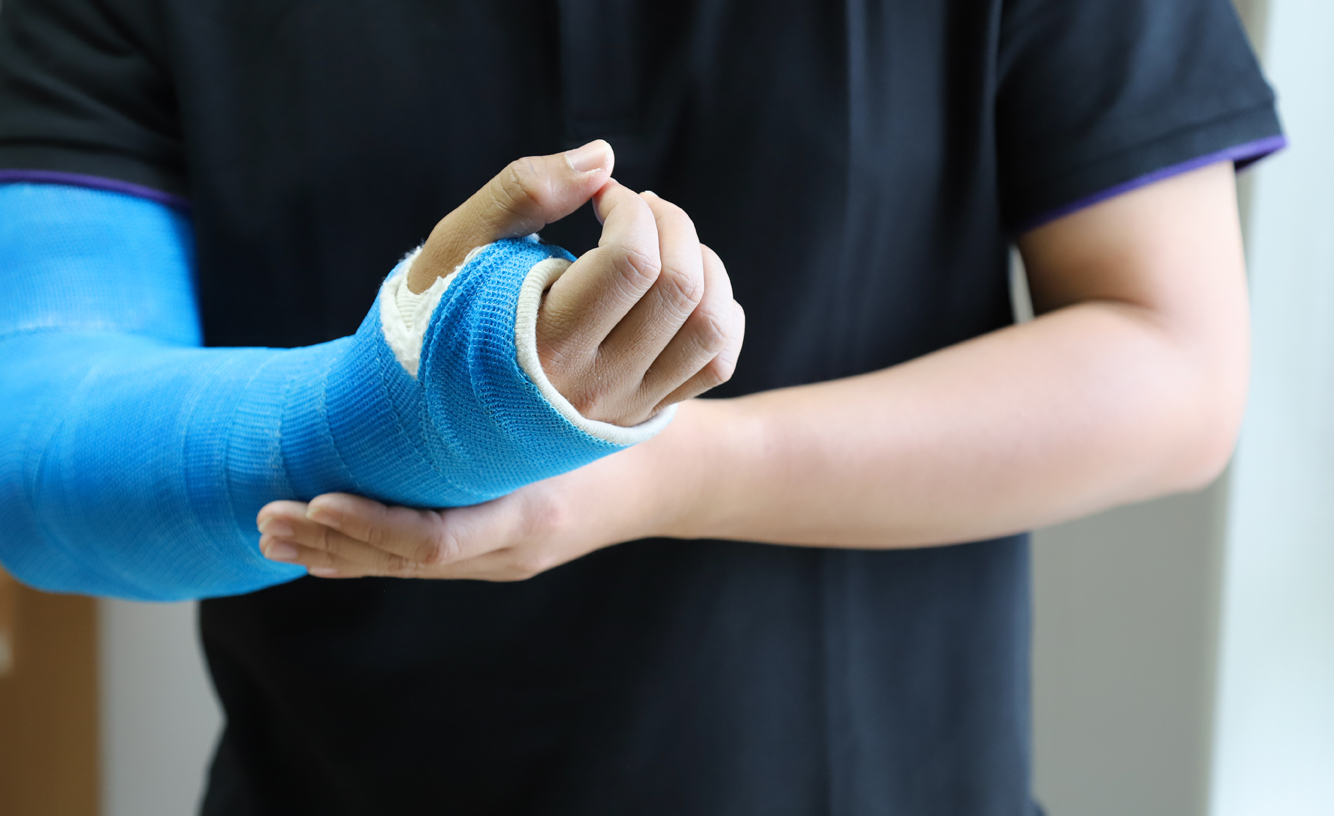}
\caption{Example of lexical ambiguity ``plaster'' in ``The doctor gave him a plaster''.}
\label{fig:018.png}
\end{figure}

Utterances were crafted to reflect as much as possible realistic spoken language, with a mean length of 8 tokens (min = 5, max = 13). Whenever possible, examples were drawn from domains pertinent to interpreting scenarios (e.g., healthcare, business meetings, public presentations). This ensures that the resulting translations approximate the kinds of challenge encountered in real communicative situations. The sentences provide no additional linguistic cues that could resolve ambiguity (as co-text sometimes does). For instance, the lexical ambiguity ``plaster'' in ``He put a plaster on it, because he cut his finger while cooking.'' could be simply resolved by the co-text ``cut a finger''.

The images associated with each utterance were chosen to contain disambiguating information relevant to the trigger. For gender resolution, images depict speakers of identifiable gender; for polysemy, they illustrate the intended sense (e.g., ``bank'' as financial institution vs. river bank) or the setting in which such an object typically comes from; and for syntactic ambiguity, they clarify structural interpretation (e.g., prepositional attachment). All images were manually checked for quality and relevance.

For \Cfour{}, we created 120 mismatched caption–utterance pairs by reassigning images across unrelated utterances. These examples allow us to probe robustness when contextual input is misleading.  Table~\ref{tab:corpus} summarizes the composition of the diagnostic corpus. We constructed 40 examples for each of the three ambiguity triggers, yielding a total of 120 items. This balanced design ensures that each category is equally represented and supports controlled comparisons across conditions.

\begin{table}[t]
\centering
\caption{Corpus composition and length statistics (tokens counted on source utterance).}
\label{tab:corpus}
\begin{tabularx}{\columnwidth}{lrrr}
\toprule
Trigger & \#Items & Mean (SD)\\
\midrule
Lexical ambiguity   & 40 & 6.85 (1.93)  \\
Gender agreement      & 40 & 8.3 (4.2)  \\
Syntactic polysemy      & 40 & 11.2 (1.9)  \\

\midrule
\textbf{Total} & 120 & 8.78 (3.39) \\
\bottomrule
\end{tabularx}
\end{table}

Annotations include source transcription, gold reference translation, associated image, and trigger category. This compact yet targeted design enables fine-grained analysis of when and how visual input contributes to translation quality.

\subsection{Evaluation}

\begin{figure}[t]
\centering
\includegraphics[width=0.9\linewidth]{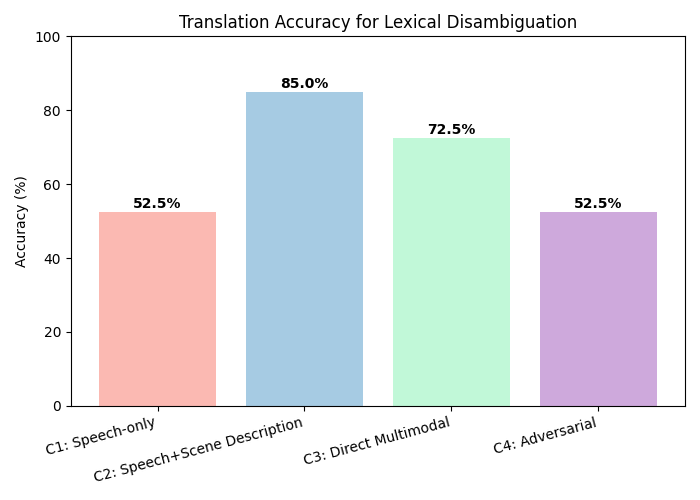}
\caption{Accuracy by condition for Lexical Ambiguity.}
\label{fig:results_lex}
\end{figure}

\paragraph{Lexical disambiguation.}  
The results show clear differences in translation
accuracy across conditions, as shown in Figure~\ref{fig:results_lex}. \Cone{} performance was 52.5\% (21/40, 95\% CI [37.5, 67.1]), not significantly different from chance ($p=.87$). \Ctwo{} yielded a strong improvement to 85.0\% (34/40, 95\% CI [71.0, 92.9]), highly above chance ($p<.001$) and significantly better than \Cone{} (McNemar $p<.002$). \Cthree{} condition also improved to 72.5\% (29/40, 95\% CI [57.2, 83.9]), significantly above chance ($p<.01$) but only marginally better than \Ctwo{} (McNemar $p\approx.065$). Performance in the \Cfour{} condition fell back to baseline (52.5\%, $p=.87$). These results demonstrate that relevant visual cues — particularly textual captions — strongly support lexical disambiguation.  

\begin{figure}[t]
\centering
\includegraphics[width=0.9\linewidth]{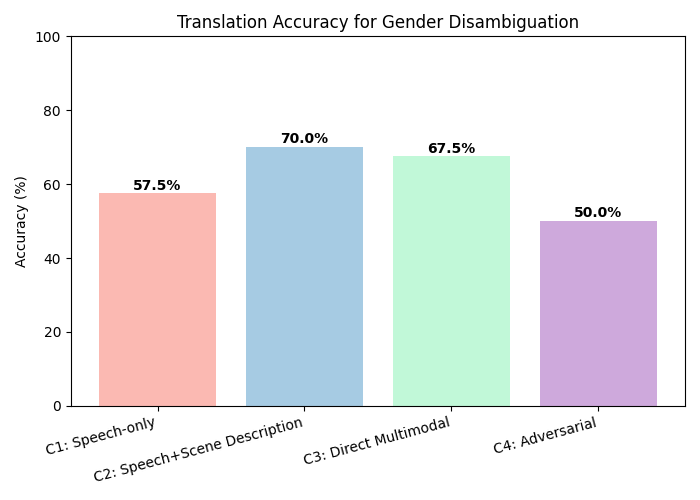}
\caption{Accuracy by condition for Gender Disambiguation.}
\label{fig:results_gender}
\end{figure}

\paragraph{Gender disambiguation.}  
For gender disambiguation (Figure~\ref{fig:results_gender}), \Ctwo{} condition reached 57.5\% (23/40, 95\% CI [42.0, 71.5]), not significantly above chance ($p=.43$). \Cthree{} increased accuracy to 70.0\% (28/40, 95\% CI [54.6, 81.9]), significantly above chance ($p=.017$) but not significantly better than \Ctwo{} (McNemar $p=.24$). \Cthree{} condition reached 67.5\% (27/40, 95\% CI [52.0, 79.9]), also above chance ($p=.038$), yet again without significant within-item improvement over \Ctwo{} (McNemar $p=.36$). The \Cfour{} condition dropped to 50.0\% (20/40, 95\% CI [35.2, 64.8]), exactly at chance ($p=1.0$). Thus, while both captions and direct multimodal provide some support, gains are weaker and less consistent than in lexical disambiguation.  
 
\begin{figure}[t]
\centering
\includegraphics[width=0.9\linewidth]{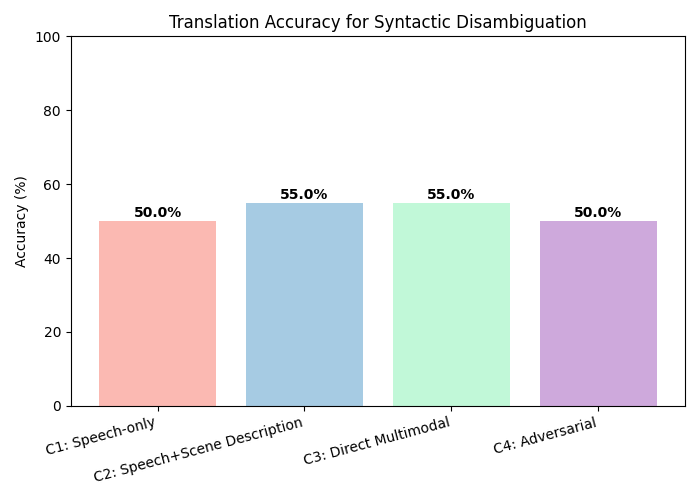}
\caption{Accuracy by condition for Syntactic Disambiguation.}
\label{fig:results_synt}
\end{figure}

\paragraph{Syntactic disambiguation.}  
For gender disambiguation (Figure~\ref{fig:results_synt}), \Cone{} condition reached 57.5\% (23/40, 95\% CI [42.0, 71.5]), not significantly above chance ($p=.43$). \Ctwo{} increased accuracy to 70.0\% (28/40, 95\% CI [54.6, 81.9]), significantly above chance ($p=.017$) but not significantly better than \Cone{} (McNemar $p=.24$). \Cthree{} condition reached 67.5\% (27/40, 95\% CI [52.0, 79.9]), also above chance ($p=.038$), yet again without significant within-item improvement over speech-only (McNemar $p=.36$). \Cfour{} condition dropped to 50.0\% (20/40, 95\% CI [35.2, 64.8]), exactly at chance ($p=1.0$). Thus, while both caption and multimodal cues provide some support, gains are weaker and less consistent than in lexical disambiguation.  
In syntactic cases, \Ctwo{} input was 50.0\% (20/40, 95\% CI [35.2, 64.8]), at chance ($p=1.0$). \Ctwo{} yielded the same result (50.0\%), and neither \Cthree{} (55.0\%, 22/40, 95\% CI [39.8, 69.3], $p=.64$) nor \Cfour{} input (55.0\%, $p=.64$) showed significant deviation from chance or improvement over \Cone{} (all McNemar $p>.65$). These findings indicate that visual and caption cues were largely ineffective for syntactic disambiguation in this dataset, contrasting with their strong impact on lexical ambiguity.  
  
\paragraph{A note on chance level.}
Each item in our test set has two equiprobable interpretations (two images designed to trigger different senses). 
Accordingly, the chance baseline is $50\%$. 
We therefore evaluate each condition not only relative to other conditions (paired McNemar tests) but also against chance using an exact binomial test with $H_0\!: p=0.5$. 
For a condition with $k$ correct out of $n$ items, we report $\hat{p}=k/n$, its $95\%$ Wilson CI, and the binomial $p$--value 
($p=\Pr\{X\le k\}\times 2$ if $k\le n/2$, else symmetric). 
Accuracies near $50\%$ indicate no effective disambiguation; values significantly above $50\%$ demonstrate that the provided cue (e.g., caption or image) adds usable information. 
When two conditions are compared on the same items, we also report McNemar’s test to quantify within-item improvements independently of base rate.

\section{Discussion}

Our evaluation compared three forms of disambiguation (lexical, gender, and syntactic) across different input modalities. Overall, the findings indicate a benefit of integrating visual cues, thereby supporting our \textbf{H1 (visual helps)} hypothesis, even if this does not applies to all tested ambiguity ty. By contrast, \textbf{H2 (caption vs.\ multimodal direct)} was not confirmed, as the two modalities performed comparably, with a slight but statistically insignificant preference for the caption-based variant. Finally, \textbf{H3 (robustness)} was confirmed: system performance did not differ between the absence of a vision component and the presence of a vision module supplying irrelevant images, suggesting that misleading visual input does not degrade performance relative to the unimodal baseline.

With respect to the different ambiguity types, the benefits were not uniform: lexical ambiguity was substantially reduced through caption concatenation and, to a lesser extent, by direct multimodal input; gender-related ambiguity showed weaker yet measurable gains; whereas syntactic ambiguity remained largely unaffected.

For lexical disambiguation, the speech-only condition performed at chance level, as expected in the absence of contextual information. The addition of structured image captions yielded a substantial and statistically robust improvement, indicating that explicit textual cues derived from visual input are highly effective for grounding lexical meaning. Direct multimodal integration also enhanced accuracy, though to a slightly lesser extent than captions, suggesting that the alignment between visual representations and linguistic disambiguation, at least in our experimental setting, remains less reliable than language-to-language mappings. Crucially, adversarial images neutralized these gains, highlighting the vulnerability of lexical disambiguation to misleading contextual input.

For gender disambiguation, improvements with both captions integration and direct multimodal input were significant relative to chance but did not consistently outperform speech-only within items. This indicates that gender information is not always reliably captured or transferred by the visual or captioning systems used here. It also points to the inherent challenge of resolving gender in translation tasks, where subtle contextual signals may be required to establish agreement or reference. The vulnerability of this task to adversarial input further highlights the fragility of current multimodal approaches in domains where social and grammatical cues interact.

Syntactic disambiguation proved the most resistant to contextual support: performance remained close to chance across all conditions, with neither captions nor multimodal input yielding measurable gains. This likely reflects the fact that the syntactic ambiguities in the dataset were designed to require structural rather than referential resolution. Visual or caption-based cues appear ill-suited to supplying the hierarchical constraints necessary for correct parsing, underscoring a limitation in how far multimodality alone can advance machine interpreting.

\section{Limitations and future work}
Beyond these benchmark findings, several system-level limitations must be acknowledged and can serve as a basis for future work. The vision module in our prototype relies on discrete sampling, which constrains its ability to capture continuous variation in the communicative setting. Continuous video analysis would provide richer support for dynamic environments by modeling temporal dependencies, event progression, and contextual continuity.\citep{islamVideoReCapRecursive2024}.

Moreover, the speech and vision streams are not semantically synchronized: when multiple speakers are visible, the system provides a general description of the scene but cannot reliably indicate, for example, who is speaking, increasing the risk of misattribution. Advances in spatial and binaural processing  offer promising avenues to mitigate this limitation \citep{chenSpatialSpeechTranslation2025}. 

Captioning quality remains a critical bottleneck. Generic or erroneous descriptions can mislead the system, as observed in the adversarial condition, or fail to provide benefits when relevant features for disambiguation are omitted. More systematic testing of prompt design is therefore warranted—for example, requesting structured or attribute-specific captions rather than general scene descriptions, as in our experiment. Our evaluation revealed several instances, particularly in gender resolution, where captions failed to encode visually available gender information. Explicitly instructing the captioning system to provide such attributes would likely improve reliability. More broadly, these findings highlight the importance of designing vision–language models for translation tasks with explicit attention to communicative functions: captions optimized for generic scene understanding may be insufficient, whereas task-specific guidance can substantially enhance their utility for machine interpreting.

Finally, our study addressed only three types of ambiguity, leaving open the question of how multimodal systems perform when confronted with broader contextual challenges such as deixis and reference, entity identification and omission, discourse-level coherence, pragmatic intent, and domain-specific knowledge.

\section{Conclusions}

We introduced the concept of \emph{Visually-Grounded Interpreting}, an extension of machine interpreting that augments the speech translation pipeline with real-time scene descriptions generated by a vision–language model. To demonstrate its feasibility, we implemented a working prototype that integrates state-of-the-art models, providing a proof of concept for the practical viability of this approach.

Using a targeted diagnostic corpus, we compared multiple integration strategies, ranging from source-only to caption-based variants, a multimodal direct setup, and an adversarial setting, on three families of ambiguity (lexical, gender, syntactic). Evaluations against a well-defined chance baseline (50\%) showed that visual grounding can materially improve translation when the ambiguity is referential: lexical disambiguation benefited substantially from visual cues. Gender disambiguation showed smaller, less consistent gains, while syntactic ambiguity remained at or near chance, indicating that the kinds of structural constraints required for syntactic resolution are not supplied by simple visual cues or generic captions.

From a system perspective, practical deployment will require addressing three limitations identified in our prototype: (i) discrete image sampling misses scene dynamics, arguing for continuous video analysis; (ii) lack of synchronization between speech and vision leads to speaker–scene misattribution in multi-person settings, motivating audiovisual alignment (e.g. spatial/binaural approaches); and (iii) caption quality remains a bottleneck because overly generic or erroneous descriptions can mislead downstream translation. 

Notwithstanding its current limitations, vision grounding emerges as a viable avenue for improving machine interpreting in ambiguity-prone cases. Its full potential, however, is likely to unfold only when combined with richer sources of context—such as discourse state, domain knowledge, and speaker metadata, and with tighter cross-modal synchronization. Future work should therefore pursue continuous, synchronized audiovisual capture in conjunction with stronger contextual modeling, and evaluate the approach in ecologically valid scenarios that better reflect the complexity of real-world interpreting. Taken together, these directions position Vision-Grounded Interpreting as a step toward next-generation speech translation systems that are more adaptive, context-sensitive, and communicatively robust.

\bibliography{paperMI}

\cleardoublepage
\appendix 






\end{document}